# Soft Fluidic Sheet Transistor for Soft Robotic System Enabling Fluid Logic Operations*

Yuki Origane, Koya Cho and Hideyuki Tsukagoshi

*Abstract*— Aiming to achieve both high functionality and flexibility in soft robot system, this paper presents a soft urethane sheet-like valve with an amplifier that can perform logical operations using only pneumatic signals. When the control chamber in the valve is pressurized, the main path is compressed along its central axis, buckling and being pressed, resulting in blockage. This allows control by a pressure signal smaller than that within the main channel. Furthermore, similar to transistors in electrical circuits, when combined, the proposed valve can perform a variety of logical operations. The basic type operates as a NOT logic element, which is named the fluidic sheet transistor (FST). By integrating multiple FSTs, logical operations such as positive logic, NAND, and NOR can be performed on a single sheet. This paper describes the operating principle, fabrication method, and characteristics of the FST, followed by a method for configuring logical operations. Moreover, we demonstrate the construction of a latch circuit (self-holding logic circuit) using FST, introducing a prototype of a fluid robot system that combines a tactile tube as a fluidic detector and fluid actuators. This demonstrates that it is possible to generate behavior that actively changes posture when hitting an obstacle using only air pressure from a single pipe, which verifies the effectiveness of the proposed methods.

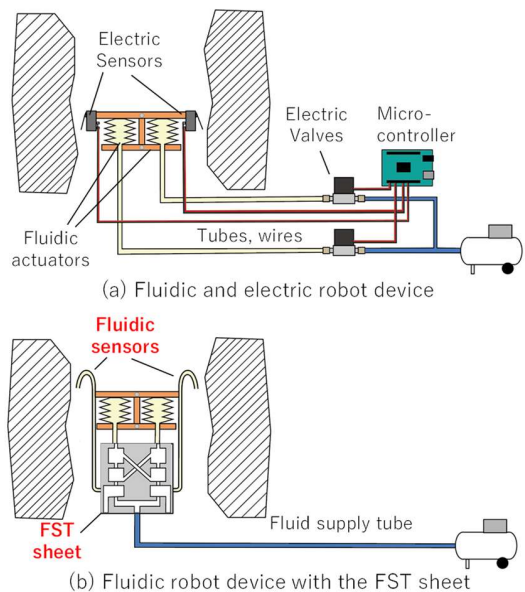

Fig. 1. Concept of robotic device with the FST

## I. INTRODUCTION

Soft robotics is a research field focused on developing robots using soft actuators made from materials such as sheets, tubes, or fabrics. This field has attracted increasing attention in recent years, particularly for applications in disaster rescue[1-4] or scenarios involving close interaction with the human body[5-10]. Among them, fluid-powered soft robots can perform a certain degree of adaptability to the external environment through passive deformation under open-loop control.

However, to achieve high-performance to more complex environments, active deformation of fluid actuators by feedback control is essential. To realize it, a control system composed of a detector, calculating unit, amplifier, and control unit is required. In the case of fluid actuator control, each of them uses a sensor, microcontroller, transistor, and solenoid valve in general (Fig.1(a)). However, this results in a configuration in which electrical and fluid signals are mixed. As a result, when attempting to build a robot with multiple degrees of freedom, the number of wiring and piping leads to increase, making the structure more complicated. If the solenoid valve is located away from the actuator, the piping will obstruct its movement. If located near the actuator, the rigid and bulky structure of the solenoid valve and the electric cable will impair the flexibility of the actuator.

One solution might be to use only fluid signals instead of electrical signals, thereby reducing the number of wiring lines. This would be especially useful in disaster areas where explosion protection is required, and in medical settings where prevention of electric shock is essential. From this point of view, Fluidics by purely fluid-control elements has been studied [11]. This element is based on the principle of the Coanda effect, in which a jet of fluid adheres to the sidewall [12]. However, because this element is delicate, a configuration without moving parts is desirable, and a rigid structure is essential. Therefore, flexibility is not realisitic. Furthermore, although this element performs the functions of a detector, a calculator, and an amplifier individually, multiple elements must be combined to realize a feedback system [13]. In addition to them, a valve must also be applied to an actuator. As a result, the entire system is bulky.

An another option without using electrical signals, several studies have reported on mechanically operated valves driven by the deformation force of flexible actuators. One example is a feedback system using a mechanical valve operated by a McKibben artificial muscle [14]. Other studies have also achieved similar control by using the bending of soft tubes [15, 16]. Other approaches include a valve that mimics the vascular closure action of biological muscles [17] and a self-oscillating valve that controls airflow using magnetic attraction [18]. These approaches have two major challenges. First, every approach relies on rigid elements, such as a casing to compress the tube or a magnet to generate force, which compromises the flexibility of the overall system. Second, achieving complex logic functions would require the combination of numerous pneumatic tubes and valves, complicating system integration.

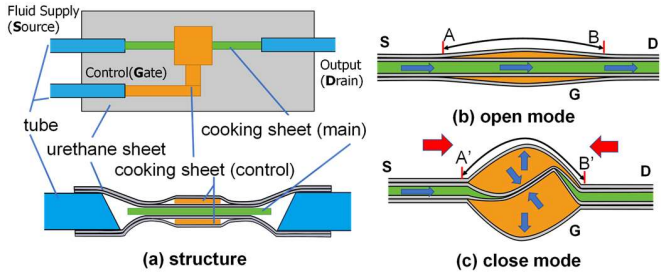

Fig. 2. Basic structure of the FST

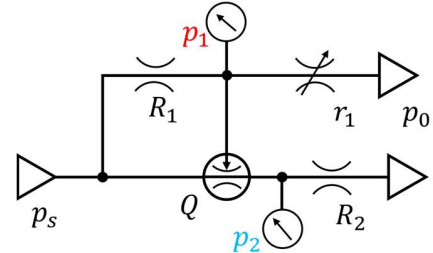

Fig. 4. Fluidic Circuit for Validation of FST

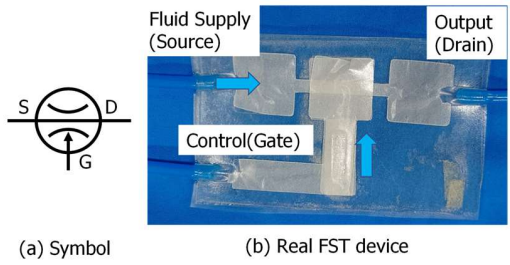

Fig. 3. Symbol and real device of the FST.

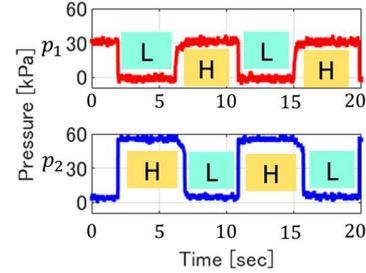

Fig. 5. Logical behavior of the single FST

To address these challenges, we propose a flexible sheet-like valve based on a urethane sheet, capable of incorporating a calculator, amplifier, and control unit. This device is named the fluidic sheet transistor (FST) because of its function to a transistor for electric circuit. A single FST functions as a NOT gate, and by combining multiple FSTs, integrated circuits that perform positive logic, NAND, and NOR logic using only air pressure signals can be constructed. FSTs have following four main advantages: 1) Transistor-like switching functions can be achieved with air pressure signals smaller than those of the main flow path. 2) In principle, any logical operation can be implemented within a single sheet by combining FSTs. 3) Consisting only of a sheet and tubes, the proposed valve can be fabricated by transfer printing onto a flexible sheet. 4) Driven solely by air pressure signals, no other supply is required (Fig.1(b)).

In Chapter 2, we introduce the operating principles, fabrication method, and flow control characteristics of basic FST devices. Chapter 3 shows how to combine multiple FSTs to build integrated logic circuits, including Positive logic, NAND gates, NOR gates. Chapter 4 shows how these circuits can be further extended to implement self-holding latches, and provides an example application of tactile tubes integrated with fluidic actuators.

## II. BASIC DESIGN OF FLUIDIC SHEET TRANSISTOR

### A. Basic Principle

First, we introduce basic mechanism of a FST, which is a fundamental unit of the fluid logic devices introduced in the later section. The FST consists four urethane sheet layers. By heat pressing after inserting heat resistant sheets like cooking sheet between urethane sheets, we can create three layers of air pocket. When fluid pressure is applied to these layers, the air pockets expand and allow the fluid to pass through. Fig. 2 shows the typical design of the flow path for the FST. Similar to electrical transistor such as field effect transistor (FET), the FST has three ports. Main flow path (green line in Fig. 2) from fluid supply port (source) to output (drain) is in the center path layer. The gate port, which connects to path for control (orange area in Fig. 2) sandwich the main path from top and bottom layer by chamber. When we cut fluid flow from gate, flow passes freely from the source port to the drain port (Fig. 2(b)). On the other hands, when we supply fluid flow to control path through the gate port, chambers expand and close main path. Since the urethane sheet is nearly inextensible, inflation of the control chamber generates tensile forces on both sides of the main path, leading to its buckling and partial obstruction of fluid flow. At the same time, the pressure within the control chamber compresses the main path directly, resulting in its complete closure as Fig. 2(c). This allows the fluid flow from drain to switch opposite side of fluid pressure to the gate port.

This behavior is similar to the P-ch FET. Both the P-channel FET and the FST open a path between the source and drain when the gate pressure decreases. However, unlike the FET, the FST has no directional preference in its main flow path, allowing fluid to flow in either direction. As a result, the drain and source are not distinguished for the FST. In the rest of the paper, we use a symbol (Fig. 3(a)) like a variable throttle valve to express the FST in flow circuits. In the symbol, horizontal line between throttle shows the main path from the source port to the drain port. Arrow from bottom shows the control path (connected to the gate port).

### B. Amplifier

To validate how large the pressure in the control chamber can control the main flow path, we investigated the real FST

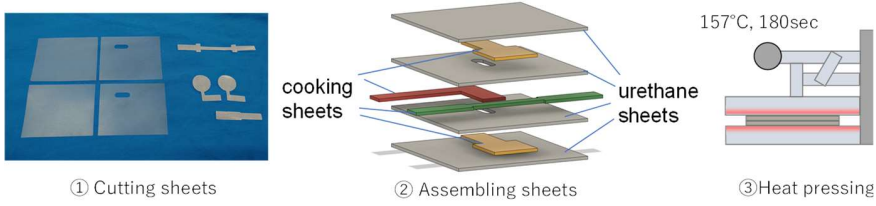

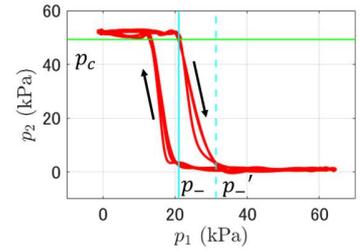

Fig. 6. Fabrication process of the FST

Fig. 7. Typical hysteresis curve of the FST

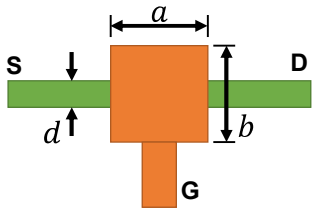

Fig. 8. Design parameters

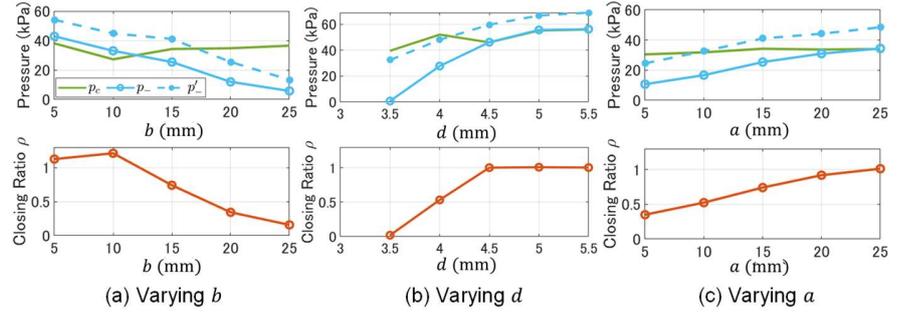

(a) Varying $b$    (b) Varying $d$    (c) Varying $a$

Fig. 9. Closure characteristic of the FST

(Fig. 3(b)). Fig. 4 shows the circuit for the validation. $p_s$ and $p_0$ denote supply pressure for the circuit and atmospheric pressure, respectively. Manual valve controls air pressure $p_1$ applied to the gate port of the FST. The drain port of the FST is connected to the atmosphere through a restrictor $R_2$, which is implemented using a thin tube. Without the restrictor, the drain pressure $p_2$ would indicate atmospheric pressure $p_0$ regardless of whether the FST is open or closed. With the presence of the restrictor $R_2$, when the FST is open, airflow occurs from the source to the drain, and this flow creates a pressure difference across $R_2$, which can be measured as a change in $p_2$.

Fig 5 shows time series data of $p_1$ (above plot with red line) and $p_2$ (bottom side plot with blue line) while we switch manual valve repeatedly. We set supply pressure as $p_s = 70$ kPa in this experiment. In Fig. 5, we can see that the $p_2$ behaves as mirror of $p_1$. These results confirm that the FST exhibits NOT logic behavior, where $p_1$ serves as the input and $p_2$ as the output, and the output logic is inverted relative to the input.

Also, the required control pressure could be achieved with half the pressure of the main path. This means that the proposed tube-compression valve possesses a function of amplifier capable of controlling a larger pressure in the main flow path with a smaller one.

Conventional pilot valves using the poppet and spool have been utilized to achieve a similar function. However, they have a rigid structure difficult to make thin, and are operated by a solenoid drive to generate the differential pressure on both sides of the valve.

The tube-compression valve proposed here is totally different in both structure and principle from these, as it is sheet-shaped and can be controlled with only a pneumatic signal, utilizing buckling to allow for small signal control.

*C. Fabrication*

In this section, we describe about fabrication process of the FST (Fig. 6). First, as shown in Fig. 6(1), the urethane sheets and the cooking sheet are cut into the desired shapes. By adjusting the shape of the cooking sheet, the flow paths and chambers can be freely arranged. Among the urethane sheets, holes are punched in the two central layers to allow air from the gate port to be supplied to both the upper and lower control chambers. Next, as illustrated in Fig. 6(2), the urethane sheets and cooking sheet are stacked alternately. The stacked layers are then placed into a heat press machine and thermally laminated at 157 °C for 3 minutes (Fig. 6(3)). Through this thermal lamination process, the urethane sheets are fused together. However, the areas where the cooking sheet is sandwiched are not bonded, forming pockets that function as flow paths and chambers. Finally, the cooking sheet at the ends is removed, and tubes are attached to supply air.

*D. Hysteresis*

Here, we denote about closing-opening hysteresis. Fig. 7 shows the one typical hysteresis curve obtained from experiment using circuit Fig. 4. The horizontal axis represents the input pressure $p_1$ applied to the gate of the FST, and the vertical axis represents the output pressure $p_2$ measured at the drain port. The hysteresis curve is traced clockwise starting from the upper left. As $p_1$ gradually increases, $p_2$ begins to decrease around $p_1 = 20$ kPa, indicating that the flow path starts to close. Subsequently, $p_2$ continues to decrease monotonically with $p_1$, following an approximately linear trend. When $p_1$ reaches around 30 kPa, the path is completely closed. After complete closure, $p_1$ is decreased. In this process, $p_2$ remains nearly constant until $p_1$ falls to around 20 kPa, indicating that the closed state is maintained. Around $p_1 = 10$ kPa, the path reopens abruptly, and the value of $p_2$ returns to its high-pressure state $p_2 = 50$ kPa.

This result shows the presence of hysteresis: the gate pressure required to close the path is higher than that required to reopen it, indicating an asymmetry in the opening and closing behavior. This hysteresis is considered to originate from the energy loss characteristics of polyurethane materials during deformation [19].

*E. Gate Control Pressure and Design Parameters*

This section describes the relationship between the design parameters of the FST and its performance. First, we define the closure start gate pressure $p_-$, closure end gate pressure $p'_-$, and blocked drain pressure $p_c$ (see also Fig. 7). We define the closure start pressure ratio ρ as follows:

$$\rho = \frac{p_-}{p_c}. \quad (1)$$

A smaller value of ρ means that a high drain pressure can be controlled with a low gate pressure, and in practical applications, an FST with a smaller ρ is preferred. As design parameters, we consider the length of horizontal direction to the main path $a$ and the vertical length $b$ of the control chamber, and the width $d$ of the main path (Fig. 8). We fabricated FSTs by changing one parameter at a time while fixing the other two, using $a = 15$ (mm), $b = 15$ (mm), and $d = 4$ (mm) as the baseline. For each fabricated FST, we experimentally measured the hysteresis curve and calculated ρ. The relationship between ρ and each parameter is shown in the Fig. 9. When $b$ is varied, as shown in the Fig. 9(a), ρ stays around 1 up to $b = 10$ mm, then decreases below 1 as $b$ increases beyond 15 mm, reaching $\rho = 0.1$ at $b = 25$ (mm). This indicates that increasing $b$ improves the closure performance. Next, Fig. 9(b) shows the closure characteristics when $d$ is varied. There is missing data for $d = 3$, because the main path was too narrow for air to pass through, making pressure measurement impossible. Notably, ρ increases significantly from $d = 3.5$ (mm) to $d = 4.5$ (mm), indicating that increasing $d$ degrades closure performance. These results imply that increasing $b$ enlarges the pressure-receiving area of the control chamber, thereby increasing the force to buckle the main path at the same gate pressure. On the other hand, increasing $d$ strengthens the main path's resistance to buckling, which in turn requires a higher gate pressure for closure.

Although $a$ also contributes to increasing the size of the control chamber, the trend in closure pressure is opposite to that of $b$. Looking at Fig. 9(c), we see that ρ is around 0.4 at $a = 5$ (mm), but increases as $a$ becomes larger. When $a$ is large, the tube bends gently even when both ends are compressed, so airflow is not blocked (Fig. 10(a)). In contrast, when $a$ is relatively small, compressing both ends of the main path by the control chamber causes a buckling with small bending radius that obstructs airflow (Fig. 10(b)). Due to this effect, reducing $a$ (within the range where airflow is still possible) decreases the gate pressure required for closure.

From the above, we conclude that to design an FST with low closure pressure, it is effective to use a rectangular control chamber with a larger dimension perpendicular to the main path and a smaller dimension parallel to it, and to make the main path narrower. To minimize the FST's footprint while keeping the closure pressure low, the main path must be narrowed. However, this increases the pressure loss (on-resistance) from source to drain, making it unsuitable for controlling devices that require high flow rates.

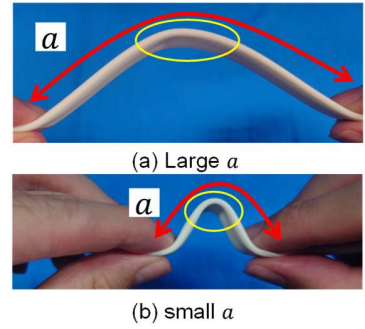

(a) Large $a$

(b) small $a$

Fig. 10. Image of buckling main path

### III. APPLICATION FOR LOGIC DEVICES

In this section, we construct basic logic devices by combining multiple FSTs. A key feature of these components is that all of them can be implemented on a single sheet.

*A. Positive Logic*

We first introduce a positive-logic switching device by cascading two FST-based NOT gates in series. Fig. 11(a)(i) and (ii) illustrate the structure of this positive-logic device. When the control A port is in the low state, the control-side FST remains open. As a result, pressure is applied to the gate of the output-side FST, causing it to close. Consequently, no pressure is delivered to port Q, which becomes low. Conversely, when the control A port is high, the control-side FST is closed, while the output-side FST remains open. In this case, pressure is supplied from the supply port to port Q, resulting in a high output. In this manner, by passing through two stages of FST-based logical inversion, we obtain a device whose output terminal reflects the same logical state as the control terminal—thus realizing a positive-logic element.

The gate of the output-side FST is connected to the atmosphere through a pneumatic resistance $R$. This allows exhaust from the gate when the control-side FST is closed, resetting the output-side FST to the low state. If this connection to the atmosphere were absent, residual air trapped in the gate region will keep the output-side FST closed regardless of the state of the control-side FST. Proper venting through independent paths is essential in the design of pneumatic logic circuits.

Figure 11(a)(iii) shows the experimental results, which confirm that the logical states of the control and output terminals match, thereby validating the logic behavior shown in (iv).

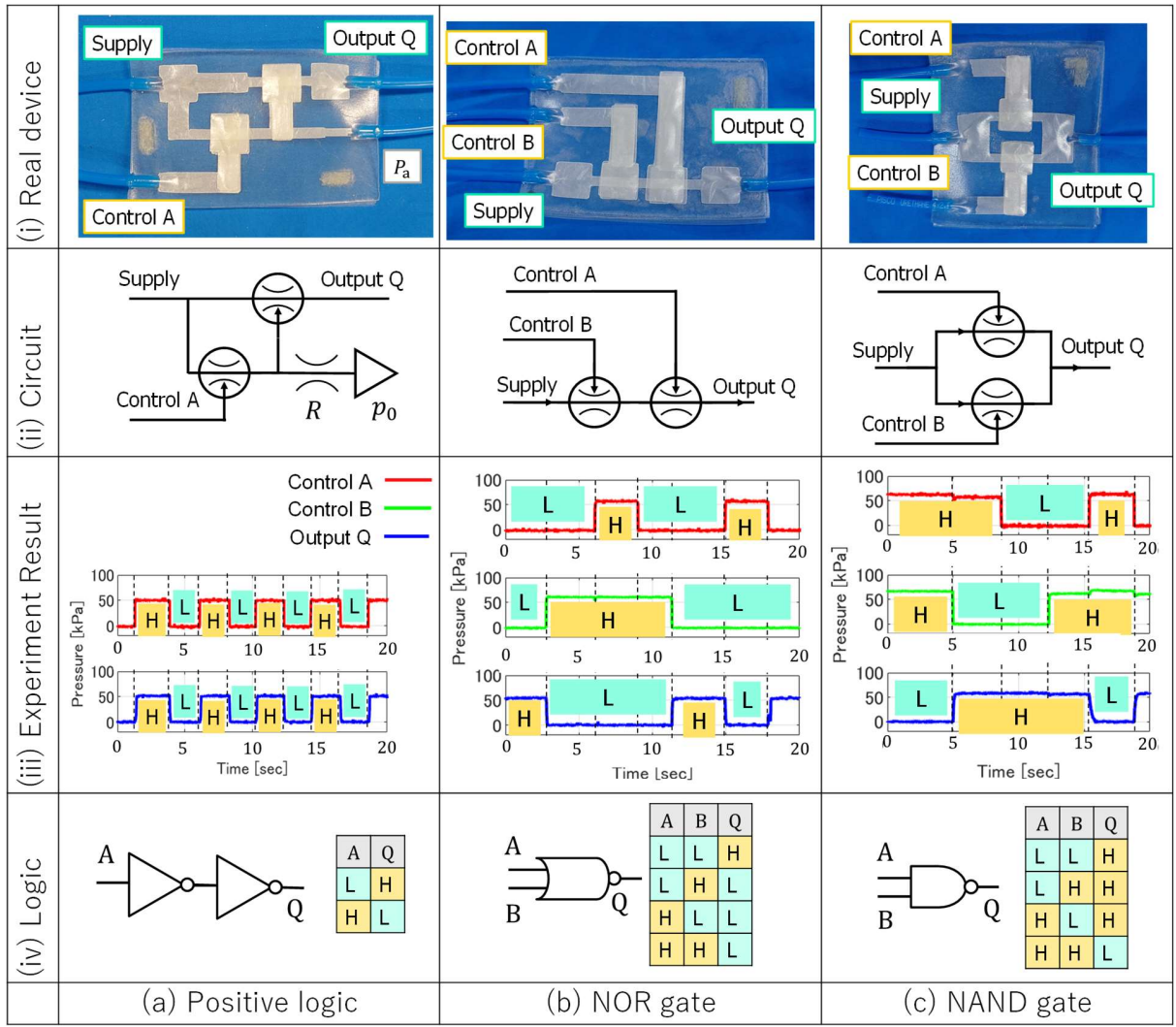

Fig. 11. Logical devices

*B. NOR and NAND gates*

NOR and NAND gates can also be implemented using two FSTs. Figures 11(b)(i) and (ii) show the structure of the NOR gate. A single flow path from the supply port to the output port is controlled by two FSTs. As illustrated in Fig. 11(b)(iii), the flow path remains open only when both control ports A and B are in the low state; in all other cases, the path is closed. This configuration enables the NOR logic behavior shown in Fig. 11(b)(iv).

In the NAND gate, there are two independent flow paths from the supply port to the output port, as shown in Figures 11(c)(i) and (ii). Each path is individually controlled by an FST, allowing the output port to remain high as long as either control port is in the low state. Figure 11(c)(iii) shows the experimental results: the Output becomes low only when both A and B ports are high, and remains high in all other cases, confirming the NAND logic behavior.

These results demonstrate that both NOR and NAND gates can be realized on a single sheet by appropriately combining multiple FSTs.

IV. APPLICATION FOR SOFT ROBOTIC DEVICE

In this section, we introduce a robotic device that integrates sensory and moving functions by combining FSTs with soft sensors and actuators, in order to demonstrate the applicability of the FST to soft robotic systems.

The behavior of switching outputs in response to sensor inputs is fundamental in robotic control systems. For instance, in a mobile robot, it may turn left when an obstacle is detected on the right, and turn right when an obstacle is detected on the left. Although this behavior is simple, it requires the ability to maintain the actuator's state after a sensor input has occurred, until the next input is received. In the field of electronic circuits, this type of self-sustaining behavior is implemented using latch circuits. A typical latch is constructed by cross-coupling the inputs and outputs of two NOR gates. As shown in the Fig. 12(iv), the latch retains the state defined by the inputs at the set and reset ports.

Fig. 12(ii) shows a latch circuit constructed using four FSTs. Fig. 12 (i) presents the actual fabricated latch sheet. The latch sheet employs four FSTs to form two NOR circuits. The two FSTs on the right provide feedback from the output of the

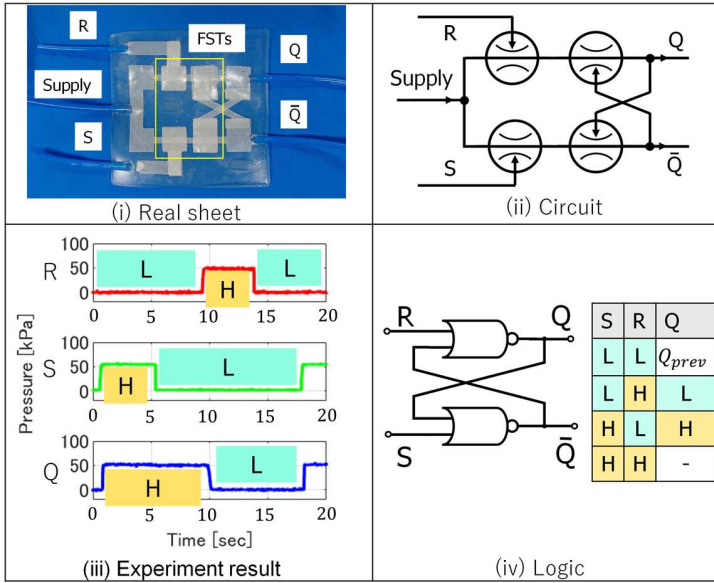
Fig. 12. Latch circuit

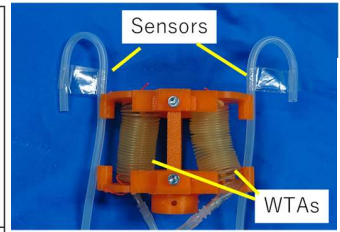
Fig. 13. Robotic device

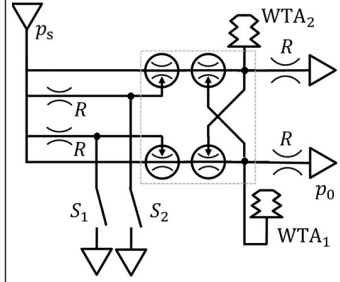
Fig. 14. Circuit of robot device with FST-used latch mechanism.

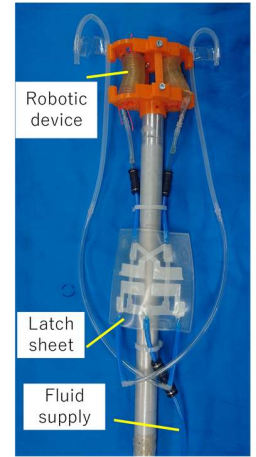
Fig. 15. Device

NOR circuit on the opposite side. Fig. 12(iii) illustrates the pressure change at output Q in response to inputs at the set and reset ports for this latch sheet. When the set port is set to high at around 0 seconds, Q also becomes high. The high state of Q is maintained even after the set port is returned to low. When the reset port is set to high at around 10 seconds, Q switches to low. This low state is also retained even after the reset port is returned to low. These results demonstrate that the logic shown in Fig. 12(iv) can be realized using the sheet composed of FSTs.

Fig. 13 illustrates the robotic device used as the control target. The system consists of two Wound Tube Actuators (WTAs [5]) and two tactile tubes as fluidic detector. Each WTA is constructed by layering flattened urethane tubes and exhibits elongation when air is supplied. When one of the WTAs extends, the orientation of the device changes accordingly. The tactile tubes were fabricated by bending silicone ones; when an object contacts them, the tube buckles, preventing air from flowing through it. By connecting the latch circuit, tactile tubes, and WTAs, we constructed the device with a circuit in the Fig. 14. The two tactile tubes, $S_1$ and $S_2$, are connected to the two inputs of the latch circuit. The outputs of the latch, $Q$ and $\bar{Q}$, are connected to the WTAs. When a tube is activated, the actuator switches its state, and in the absence of further input, the actuator retains its state due to the latch behavior. Fig. 15 illustrates the connection between the device and the sheet. The sensors and WTAs of the device are each connected to the sheet. The only external connection is a single tube that supplies pneumatic pressure.

Fig. 16 shows the behavior of the robotic device. In this experiment, the device was moved in the horizontal direction manually. At time 0 s, the device is facing to the left. After the left tube contacts an obstacle at time 1.0 s, the left-side WTA inflates, and by time 1.5 s, the device has turned to the right. Then, when the device contacts an obstacle on the right side at time 2.0 s, the left WTA deflates while the right WTA inflates simultaneously, causing the device to turn back to the left. In this way, by using a latch circuit based on an FST, we successfully changed the direction of the device solely through pneumatic control in response to obstacle contact.

By combining this device with a locomotion actuator, it is possible to realize a mobile robot that can navigate obstacle-filled environments using only pneumatic systems. These results demonstrate that robot control can be achieved purely pneumatically, using logic circuits constructed with FSTs.

V. CONCLUSION

In this study, we proposed a fluidic sheet transistor (FST) with a calculating unit, amplifier, and controller, aiming to achieve soft robots fully controlled by pneumatics. We then discussed the operating principles, fabrication method, and characteristics. The FST is constructed by stacking urethane sheets. Expanding the control chamber compresses and buckles the main flow channel, performing a NOT operation, which enables to manipulate large pressures with small pressure signals.

By designing specific flow path patterns, multiple FSTs can be integrated into a single sheet, enabling the construction of logic elements such as NOR and NAND gates. Finally, we constructed a latch circuit inspired by electrical logic, using FSTs. By integrating this circuit with silicone tube-based sensors and WTAs, we demonstrated a robot device capable of changing its orientation in response to obstacle detection—using only pneumatic control.

To further advance the application of FSTs in soft robotics, miniaturization will be essential. For this purpose, quantitative model analysis of design parameters will be required. Future work will also involve closer integration with sensors and actuators to realize fully soft robotic systems.

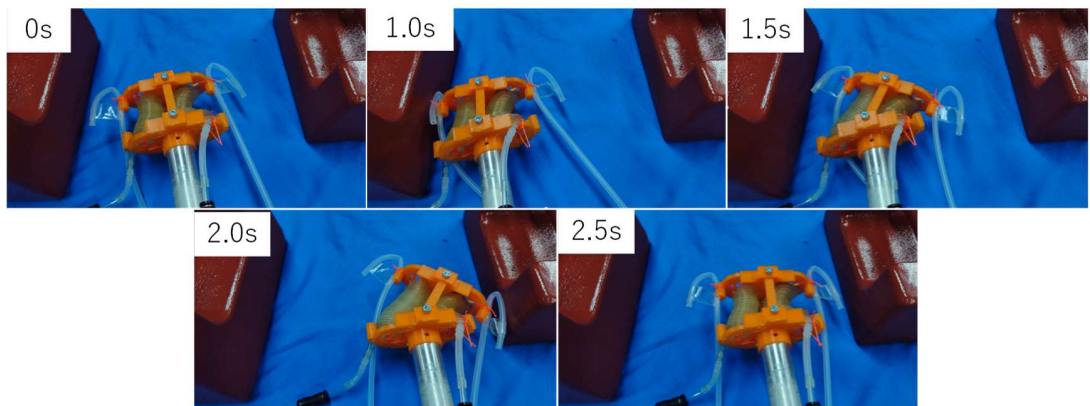

Fig. 16. Behavior of the robotic device